\documentclass[sigconf]{acmart}

\AtBeginDocument{%
  \providecommand\BibTeX{{%
    \normalfont B\kern-0.5em{\scshape i\kern-0.25em b}\kern-0.8em\TeX}}}

\copyrightyear{2021} 
\acmYear{2021} 
\setcopyright{acmlicensed}\acmConference[AIES '21]{Proceedings of the 2021 AAAI/ACM Conference on AI, Ethics, and Society}{May 19--21, 2021}{Virtual Event, USA}
\acmBooktitle{Proceedings of the 2021 AAAI/ACM Conference on AI, Ethics, and Society (AIES '21), May 19--21, 2021, Virtual Event, USA}
\acmPrice{15.00}
\acmDOI{10.1145/3461702.3462611}
\acmISBN{978-1-4503-8473-5/21/05}

\usepackage{braket}
\newcommand{\R}{\mathbb{R}}
\newcommand{\Cx}{\mathbb{C}}

\newcommand{\ketpsi}{\ket{\psi}}
\newcommand{\kz}{\ket{0}}
\newcommand{\ko}{\ket{1}}

\newcommand{\bellzz}{\frac{\ket{00} + \ket{11}}{\sqrt{2}}}

\begin{document}
\fancyhead{}

\title{Quantum Fair Machine Learning}

\author{Elija Perrier}
\email{elija.t.perrier@student.uts.edu.au}
\orcid{0000-0002-6052-6798}
% \author{Elija Perrier}
%\email{webmaster@marysville-ohio.com}
\affiliation{%
  \institution{Centre for Quantum Software \& Information,  University of Technology, Sydney}
  \streetaddress{15 Broadway}
  \city{Sydney}
  \state{NSW}
  \country{Australia}
  \postcode{2000}
}

\affiliation{%
  \institution{Humanising Machine Intelligence,  Australian National University}
%   \streetaddress{[]}
  \city{Acton}
  \state{ACT}
  \country{Australia}
  \postcode{2601}
}

% \affiliation{%

% \affiliation{%

% \affiliation{%

% \affiliation{%

% \affiliation{%

% \affiliation{%

% \affiliation{%

\renewcommand{\shortauthors}{Perrier, E.}

\begin{abstract}
  In this paper, we inaugurate the field of quantum fair machine learning. We undertake a comparative analysis of differences and similarities between classical and quantum fair machine learning algorithms, specifying how the unique features of quantum computation alter measures, metrics and remediation strategies when quantum algorithms are subject to fairness constraints. We present the first results in quantum fair machine learning by demonstrating the use of Grover's search algorithm to satisfy statistical parity constraints imposed on quantum algorithms. We provide lower-bounds on iterations needed to achieve such statistical parity within $\epsilon$-tolerance. We extend canonical Lipschitz-conditioned individual fairness criteria to the quantum setting using quantum metrics. We examine the consequences for typical measures of fairness in machine learning context when quantum information processing and quantum data are involved. Finally, we propose open questions and research programmes for this new field of interest to researchers in computer science, ethics and quantum computation.
\end{abstract}




\maketitle

\section{Introduction}

% \noindent 
Quantum computing and machine learning represent two of the most significant fields of computational science to have emerged over the last half century. Over the last decade, attention has in particular turned to the ethical implications of machine learning technology \cite{caton_fairness_2020, chouldechova_frontiers_2018, dwork_fairness_2012}, motivating the development of fair machine learning. Despite significant advances in quantum computing and quantum information processing \cite{boixo_characterizing_2018} and the growing importance of quantum devices, little or no direct work on the ethics of quantum computing has been undertaken, with limited discussion on the morality and philosophy of quantum computing \cite{aaronson_opinion_2019, aaronson_why_2011} or debates about the appropriateness of the term `quantum supremacy' \cite{palacios-berraquero_instead_2019}. 

Although fully scalable fault-tolerant \cite{preskill_fault-tolerant_1997} quantum computing has not yet been realised, the availability of intermediate noisy scalable quantum computing devices means that near-term computation on quantum devices using quantum algorithms is becoming a reality. The potential capacity of quantum computers to (i) solve problems that are classically intractable and, more practically, to (ii) solve problems which, while tractable, are classically \textit{infeasible} (due to resource constraints), motivates their use in machine learning. The wide scope of tasks that may be performed on quantum computers such as solving optimisation problems, simulating classical or complex systems or modelling highly complex systems, such as multi-agent interactive systems means that machine learning (and computation) involving quantum computers is likely to give rise to (and be subject to) ethical considerations. Many such problems remain infeasible (if tractable) for classical computation and are thus ethically unproblematic. Yet such computations may become feasible due to the enhanced computational potential afforded by quantum computing, which expands the frontiers of optimisation problems that may be solved by providing for computational capacity beyond the constraints of classical computation. If optimisation problems that are classically infeasible become computationally feasible on a quantum computer, then the ethics of such computations become relevant \textit{and} such ethics must be considered in light of the quantum nature of computation. Thus research into fair machine learning and fair computation on quantum computers is well-motivated.

Fair machine learning is the most mature field addressing the theoretical and technical challenges of undertaking classical computation subject to ethical (fairness) constraints. To date, however, no equivalent research has been undertaken regarding the technical impact of ethical constraints upon quantum computation. In this paper, we seek to address this gap via an exposition of quantum fair machine learning (see also concurrent work on ethical quantum computing generally \cite{perrier_elija_ethical_2021}). As we detail below, the unique aspects of \textit{quantum} computation (including its inherently measurement-based probabilistic nature, the availability of coherent superposition states, the restriction of computational dynamics to unitary evolution, no cloning theorems etc) mean that computation is undertaken differently from the classical case. As a result, the types of metrics, measures, circuits and ways in which constrained optimisation for fair objectives can be undertaken on a quantum computer exhibit important differences. 

For example, in contrast to classical data, quantum data is represented as vectors on Hilbert spaces $\mathcal{H}$. Classical data can in theory be directly accessed and operated upon. Quantum data, in a superposition state, cannot be and instead must be manipulated indirectly. As distinct from classical computing dynamics, quantum computing must evolve unitarily according to Schr{\"o}dinger's equation. Instead of direct probabilities, quantum states are characterised by complex-valued amplitudes which determine the probability of obtaining an outcome when measuring a quantum state $\ketpsi$. Unlike classical computing, the \textit{order} of operations matters in the quantum context, as represented by non-commutative operators (or Lie brackets) $[A,B]\neq 0$. Distance measures in quantum computing also differ. Quantum computing also has available to it important non-classical resources, such as entanglement, with no counterpart classically. As such, the study and practical implementation of fair quantum machine learning requires researchers to understand the implications of these computational differences and, in turn how these differences affect (a) the applicability of fair machine learning techniques to fair quantum computation and (b) the need to develop new quantum-specific techniques to solve ethical quantum optimisation problems.

\subsection{Results and contributions} The results and contributions of this paper are as follows.
\begin{enumerate}
    \item We present the first results characterising ways in which fair machine learning undertaken on quantum devices differs from or is similar to its classical counterpart.
    \item We provide quantum analogues of a number of key results in fair machine learning at the preprocessing, in-processing and post-processing stages, such as fairness criteria and measures and remedial strategies to satisfy fairness constraints. 
    \item We present the first technical results in quantum fair machine learning, demonstrating the use of Grover's quantum search algorithm to enable quantum algorithms achieve statistical parity within $\epsilon$-tolerance.
    \item We present the quantum analogue of Lipschitz-conditioned individual fairness criteria, expressed in terms of quantum metrics and unitary operators on quantum vectors in Hilbert space.
    \item We list open questions for researchers in quantum information processing, machine learning and ethics \cite{perrier_elija_ethical_2021} to consider as part of their research programmes.
\end{enumerate}

\subsection{Structure} The structure of the paper is as follows. \textbf{Part I} summarises the key characteristics of quantum information processing of relevance to fair machine learning, including quantum postulates, density matrix formalism, measurement statistics, no cloning theorems and open systems. In general, the interface of quantum and classical computing is usefully categorised according to the quantum-classical quadrant in which \textit{data} and \textit{processing} are categorised as either quantum ($Q$) or classical ($C$) \cite{aimeur_machine_2006, schuld_supervised_2018}. We focus upon the $CQ$ case, where classical data inputs subject to quantum information processing. We review methods by which classical data is encoded in quantum states $\ketpsi$ using basis, amplitude, sample and dynamic encoding strategies. We examine the use of quantum algorithms to solve optimisation problems subject to fairness constraints and the impact that the specifically quantum aspects of those computations has (if any) upon typical fair machine learning fairness criteria. In \textbf{Part II}, we present a comparative analysis of how typical fair machine learning characteristics differ when computation is undertaken on quantum devices. We explicitly examine quantum analogues of a number of key features of FML algorithms, including group-based fairness measures (parity, confusion matrix metric, calibration and others), individual fairness measures (e.g. Lipschitz conditioned \cite{dwork_fairness_2012} problems) and remedial strategies (such as redaction, blinding, sampling, data transformation, relabelling, perturbation and constraint optimisation). We discuss how quantum information processing renders existing techniques in FML the same, different and in need of modification. We discuss types of uniquely quantum fair computation and present the quantum analogue of Lipschitz-conditioned individual fairness constraints for machine learning algorithms. In \textbf{Part III}, we present an explicit example of how to use an adaptation of Grover's infamous quantum searching algorithm \cite{grover_fast_1996} (amplitude amplification) to satisfy fair statistical parity criteria and give a lower bound on the number of applications of unitary algorithms necessary for doing so.  In \textbf{Part IV}, we discuss open questions and provide suggestions for research programmes in quantum FML. 

% The motivation for understanding quantum fair machine learning is twofold (a) the potential for quantum computers to render computationally feasible certain ethical problems currently beyond the scope of classical computation (e.g. potentially solving higher-order complexity class problems such as multi-agent games in an ethical setting to model societal behaviour); and (b) 

\section{Part I: Quantum information processing}
In this section, we review the key distinguishing characteristics of quantum information as distinct from its classical counterpart of relevance to canonical problems in fair machine learning. We focus on aspects of quantum information processing, such as superposition states, quantum entanglement, the inherently probabilistic and measurement-based nature of quantum computing of relevance to understanding fair machine learning in a quantum context. We begin with a synopsis of the primary quantum postulates. 

\subsection{Postulates}
Quantum information processing is characterised by constraints upon how information is represented and processed arising from the foundational postulates of quantum mechanics, which we set-out as follows \cite{nielsen_quantum_2010}:
\begin{enumerate}
    \item \textit{State space}: quantum systems are completely described by (unit) state vectors within a complex-valued vector space Hilbert space $\mathcal{H}$. Typically for our formulation, we will deal with arbitrary two-level quantum systems (two-dimensional state spaces) of qubits with arbitrary state vectors with orthonormal bases $\{\kz,\ko\}$:
    \begin{align}
        \ketpsi = a\kz + b\ko
        \label{eqn:qubit}
    \end{align}
    with usual normalisation conditions on unit vectors $\braket{\psi | \psi} = 1$ (that is $|a|^2 + |b|^2=1)$, where $a,b \in \Cx$ are amplitudes for measuring outcomes of $\kz,\ko$ respectively (where $\braket{\psi | \psi'}$ denotes the inner product of quantum states $\ketpsi,\ket{\psi'}$). In density operator formalism, the system is described via the positive density operator $\rho$ with trace unity acting on the state space of the system (such that if the system is in state $\rho_i$ with probability $p_i$ then $\rho = \sum_i p_i \rho_i$). In this work, we assume the standard orthonormal computational basis $\{\kz,\ko\}$ such that $\braket{1 | 0}=\braket{0 | 1}=0$ and $\braket{1|1}=\braket{0|0}=1$.
    \item \textit{Evolution}: closed quantum systems (which we focus on in this paper for simplicity) evolve over time $\Delta t= t_1 - t_0$ via unitary transformations $U(t)=\exp(-iH(t))$ i.e. where such unitaries represent solutions to the time-dependent Schr{\"o}dinger equation governing evolution:
    \begin{align}
        i\hbar \frac{d\ket{\psi(t)}}{dt}=H(t)\ket{\psi(t)}
    \end{align} 
    where $\hbar$ is set to unity for convenience and $H(t)$ represents the linear Hermitian operator (Hamiltonian) of the closed system. The dynamics of the quantum system are completely described by the Hamiltonian operator acting on the state $\ketpsi$ such that $\ket{\psi(t)} = U(t)\ket{\psi(t=0)}$. In density operator notation, this is represented as $\rho(t) = U(t) \rho(t_0) U(t)^\dagger$. Typically solving the continuous form of the Schr{\"o}dinger equation is intractable or infeasible, so a discretised approximation as a discrete quantum circuit (where each gate $U_i$ is run for a sufficiently small time-step $\Delta$t) is used (e.g. via Trotter-Suzuki decompositions).\\
    \\
    The Hamiltonian $H(t)$ of a system is the most important tool for mathematically characterising dynamics of a system, encoding the computational processing of data encoded into quantum states and specifying how the quantum computation may be controlled. To the extent that computational processes or outcomes are important to ethical criteria (for example, understanding the dynamics giving rise to bias or discriminatory outcomes, or ensuring fair representation learning in a quantum setting), they must usually be encoded in the Hamiltonian of quantum systems. Unitary evolution is a requirement to preserve quantum coherence and probability (which give rise to the enhanced computational power of quantum systems). Thus modifications to computational subroutines in quantum computing must be done indirectly, via adjustments to the Hamiltonian which steer quantum systems towards target states using unitary evolution. 
    \item \textit{Measurement}: quantum measurements are framed as sets of measurement operators $\{ M_m\}$, where $m$ indexes the outcome of a measurement (e.g. an energy level or state indicator). The probability $p(m)$ of outcome $m$ upon measuring $\ketpsi$ is represented by such operators acting on the state such that $p(m) = \braket{\psi | M_m^\dagger M_m |\psi}$ (alternatively, $p(m) = \text{tr}(M_m^\dagger M_m \rho)$) with the post-measurement state $\ket{\psi'}$ given by: 
    \begin{align}
        \ket{\psi'} = \frac{M_m \ketpsi}{\sqrt{\braket{\psi | M_m^\dagger M_m | \psi}}}
        \label{eqn:postmeasurementstate}
    \end{align}
    The set of measurement operators $\sum_m M_m^\dagger M_m = I$, reflecting the probabilistic nature of measurement outcomes. In more advanced treatments, POVM formalism more fully describes the measurement statistics and post-measurement state of the system. There we define a set of positive operators $\{  E_m \}=\{M^\dagger_m M_m\}$ satisfying $\sum_m E_m=\mathbb{I}$ in a way that gives us a complete set of positive operators (such formalism being more general than simply relying on projection operators). As we are interested in probability distributions rather than individual probabilities from a single measurement, we calculate the probability distribution over outcomes via Born rule using the trace $p(E_i) = \text{tr}(E_i \rho)$. This formulation is important to quantum fairness metrics discussed below.

    \item \textit{Composite systems}: states $\ketpsi$ in the Hilbert space may be composite systems, described as the tensor product of states spaces of the component physical systems, that is $\ketpsi = \otimes_i \ket{\psi_i}$. We also mention here the importance of open quantum systems where a total system Hamiltonian $H$ can be decomposed as $H = H_S + H_E + H_I$, comprising a closed quantum system Hamiltonian $H_S$, an environment Hamiltonian $H_E$ an interaction Hamiltonian term $H_I$, which is typically how noise is modelled in quantum contexts. Open quantum systems are modelled usually by master equations (beyond the scope of this introductory paper), but they are relevant to quantum fair machine learning in particular because modelling the dissipative effects of system/environment interaction, or engineering dissipation so as to simulate dissipative characteristics of neural networks (see \cite{schuld_supervised_2018}) will require in certain contexts open systems' formulations. 
\end{enumerate}
Other key concepts necessary to understand the formalism below include: (a) \textit{relative phase}, that for a qubit system, where amplitudes $a$ and $b$ differ by a relative phase if $a = \exp(i\theta)b, \theta \in \R$ (as discussed below, classical information is typically encoded in both basis states e.g. $\kz,\ko$ and in relative phases); (b) \textit{entanglement}, certain composite states (known as EPR or Bell states), may be entangled. For example, for a two-qubit state: 
\begin{align}
    \ketpsi = \bellzz
\end{align}
measurement of $0$ on the first qubit necessarily means that a measurement of the second qubit will result in the post-measurement state $\kz$ also i.e. the measurement statistics of each qubit correlate. Entangled states cannot be written as tensor products of component states i.e. $\ketpsi \neq \ket{\psi_1}\ket{\psi_2}$; (c) \textit{expectation}, expectation values of an operator $A$ (e.g. a measurement) can be written as $E(A) = \text{tr}(\rho A)$; (d) \textit{mixed} and \textit{pure} states, quantum systems whose states are exactly known to be $\ketpsi$, i.e. where $\psi = \ketpsi\bra{\psi}$ are \textit{pure states}, while where there is (epistemic) uncertainty about which state the system is sin, we denote it a mixed state i.e. $\rho = \sum_i p_i \rho_i$ where $\text{tr}(\rho^2)< 1$ (as all $p_i < 1$); (e) \textit{commutativity}, where two measurements are performed on a system, the outcome, unlike in the classical case, will be order-dependent if they do not commute, that is, if $[A,B]\neq 0$; and (f) \textit{no cloning}, unlike classical data, quantum data cannot be copied (for to do so requires measurement which collapses the state destroying the coherent superpositions that encode information in amplitudes). We omit a vast universe of other characteristics of relevance to quantum fair machine learning, including error-correcting codes (encoding mechanisms designed to limit or self-correct errors to achieve fault-tolerant quantum computing) which, while relevant, are beyond the scope of this paper.

\subsection{Quantum metrics.}
Metrics play a central technical role in fair machine learning, fundamentally being the basis upon which technical definitions of fairness are constructed. Metrics for quantum information processing are related but distinct from their classical counterparts. This means translating fair machine learning metrics to the quantum realm requires different (albeit related) metric formalism. As discussed in \cite{caton_fairness_2020,corbett-davies_measure_2018, del_barrio_review_2020}, classical fairness metrics vary according to particular objectives, optimisation aims and data sources. Metrics and distance measures are an important feature of quantum information processing, with a variety of metrics deployed depending on context. For a classical bit string, there are a variety of classical information distance metrics used in general \cite{nielsen_quantum_2010}. For fair machine learning in quantum contexts, where the similarity of either input or outputs to a function is important for fairness measures (such as for Lipschitz-conditioned fairness measures below), quantum generalisations of classical distance are applicable. As information is encoded in quantum states, quantum measures of distance are therefore applicable. 
\begin{enumerate}
    \item \textit{Hamming distance}, the number of places at which two bit strings are unequal. Hamming distance is important in error-correcting contexts (of relevance to FML as quantum FML will require encoding of data in codes to achieve fault tolerance).
    \item \textit{Trace distance} or $L1$-\textit{Kolmogorov distance}, where, given two probability distributions $\{p_x \},\{q_x \}$ we have $D(p_x,q_x)=\frac{1}{2}\sum_x|p_x - q_x|$ (used in \cite{dwork_fairness_2012} a valid distance due to satisfaction of metric axioms. For quantum states represented by density matrices $\rho,\sigma$ \cite{nielsen_quantum_2010}, their trace distance is given by:
\begin{align}
    D(\rho,\sigma) = \frac{1}{2} \text{tr}|\rho - \sigma|
\end{align}
where $|\rho| = \sqrt{\rho^\dagger \rho}$ is taken as the positive square root. Trace distance has the benefit of being preserved under unitary transformations.
       \item \textit{Fidelity}, which is a common measure to assess state or operator similarity and is given by $F(\rho,\sigma) = \text{tr}\sqrt{\rho^{1/2}\sigma \rho^{1/2}}$. Along with trace distance, it is among the most important metric for similarity in quantum computing. Fidelity can be interpreted as a metric via, for example, calculating the angle $\zeta=\arccos F(\rho,\sigma)$. It is related to trace-distance via $D(\rho,\sigma) = \sqrt{1 - F(\rho,\sigma)^2}$.
    \item \textit{quantum relative entropy}, which is the quantum analogue of Shannon entropy, and is given by $S(\rho) = -\text{tr}(\rho \log \rho)$ with the quantum analogue of (binary) cross-entropy given by:
    \begin{align}
        S(\rho||\sigma) = \text{tr}(\rho \log \rho) - \text{tr}(\rho \log \sigma)
    \end{align} Both measures provide a further basis for similarity metrics required to determine if fairness criteria are met in the quantum context. 
\end{enumerate}

\subsection{Encoding data in quantum systems}
A first step in applying quantum algorithms to solve optimisation problems involving quantum data involves encoding such classical data into quantum systems in a process known as \textit{state preparation}. Encoding data fits within the preprocessing stage of typical FML taxonomies.  Standard encoding methods include \cite{schuld_supervised_2018}: (a) basis encoding, (b) amplitude encoding, (c) qsample encoding and (d) dynamic encoding. \textit{Basis encoding} encodes classical information, typically involves transforming data into classical binary bit-strings $(x_1,...,x_d), b_i \in \{0,1\}$ then mapping each bit string to the basis state of a set of qubits of a composite system. For example, for $x \in \R^N$, say a set of decimals, is converted into a $d$-dimensional bit string (e.g. $0.1\to 00001..., -0.6 \to 11001..$) suitably normalised such that $x = \sum_k^d (1/2^k)x_k$. The sequence $x$ is then represented via $\ketpsi = \ket{000001\,11001}$ (see \cite{schuld_supervised_2018}). \textit{Amplitude encoding} associates normalised classical information e.g. for an $n$-qubit system (with $2^n$ different possible (basis) states $\ket{j}$), a normalised classical sequence $x\in \Cx^{2^n}, \sum_k |x_k|^2=1$ (possibly with only real parts) with quantum amplitudes $x=(x_1,...,x_{2^n})$ can be encoded as $\ket{\psi_x}=\sum_j^{2^n} x_j \ket{j}$. Other examples of sample-based encoding (e.g. \textit{Qsample} and \textit{dynamic} encoding are also relevant but not addressed here. From a classical machine learning perspective, such encoding regimes also enable both features and labels to be encoded into quantum systems.

% One of the attractions of quantum algorithms for solving difficult or classically intractable problems lies in leveraging the uniquely quantum characteristics of quantum data. The ability to encode data in quantum superpositions, where in some sense all possible outcomes can be computed upon at once or `in parallel' (or rather, in a way that leverages the constructive and destructive interference of amplitudes). An example which we explore in Part III below is the Grover-style search algorithm based upon \textit{amplitude amplification} (speeding up search by $\sqrt{N}$ in the best case by comparison with classical counterparts).

\subsection{Quantum measurement and learning}
\paragraph{Quantum measurement}. Quantum measurement is integral to any program of ethical quantum computation. Quantum systems cannot be accessed directly. Instead, information about quantum systems depends on measurements whose outcomes are probability distributions over measurement results.  For machine learning contexts, the first key difference is that measurement outcomes in quantum computation are inherently \textit{probabilistic}. As distinct from classical computing whereby stochastic processes may be simulated, quantum computation is \textit{necessarily} stochastic, such that measurement outcomes are represented as probability distributions obtained using POVMs. Secondly, one usually only ever can sample a subspace of $\mathcal{H}$, so full probability distributions for higher-order quantum systems may be unobtainable, adding additional uncertainty.  Thirdly, where basis states are not orthogonal, then it can be proven that those states cannot be reliably distinguished (see \cite{nielsen_quantum_2010}), which impacts for example the ability to distinguish states in a way necessary to measure their metric distance, thus whether fairness constraints have been met.
% \textit{Quantum gates}. The composition of quantum algorithms can differ depending on the quadrant (CC, CQ, QC, QQ) of interest. Purely quantum algorithms are characterised as sequences of unitary operators drawn from a finite set of universal quantum gates, gates which in theory may be arbitrarily combined to perform any computation on qubits (the quantum analogue of complete logical circuit operators). For qubit systems, typical universal gate sets include Clifford group operators (permutations of the set of Pauli operators $X,Y,Z,I$) together with supplementary gates such as [T-gates] or [Toffoli] gates.

\textit{Quantum learning}. A second distinct feature of quantum algorithms is in the nature of quantum \textit{learning}. The ways in which one can learn about quantum systems and deploy learning strategies as part of measurements on or queries of quantum systems. One of the challenges of mapping classical FML to quantum computing arises in machine learning contexts. The nature of quantum information (including classical information encoded in quantum states) means that typical learning protocols, such as backpropagation, are not directly applicable because quantum information cannot be updated directly in the way that say a neural network weight operator can be. For example, the use of classical backpropagation algorithms requires classical parameterisation, say of a unitary $U(\theta)$ (where $\theta \in \R$), where online- or offline-feedback between the outcomes of measurements and classical inputs that evolve those parameters, thus the unitaries and thus the quantum states of the system, is implemented \cite{schuld_supervised_2018,youssry_beyond_2020}. More broadly, quantum algorithm formalism is usually concerned with finding optimal quantum circuits to solve an objective, however the means by which algorithms or quantum systems `learn' or `update' differs (for example, see quantum annealling-based QAOA \cite{farhi_quantum_2000}). In this paper, we include such quantum learning algorithms within the meaning of QFML, though their theoretical and implementation details will differ from hybridised classical-quantum circuits.

\section{Part II: Classical fairness measures in quantum contexts}
In this section, we compare how the unique features of quantum computation affect the applicability of techniques from classical fair machine learning to quantum settings. For convenience, we adopt the taxonomy set-out in recent reviews by Caton et al. \cite{caton_fairness_2020} and del Barrio et al. \cite{del_barrio_review_2020}. 
% FML classifications typically involve three abstract fair classification criteria, (a) \textit{independence}, where the classification (stochastic or deterministic) $R$ is independent of (protected subgroup) set membership $S$; (b) \textit{separation}, which considers the independence of $R$ and membership of $S$ conditional upon target variable $Y$ (related to equalised odds and equal opportunity metrics, see below); and (c) \textit{sufficiency}, which examines the independence of target variables $Y$ and membership of $S$ conditional upon the score $R$ (related to calibration metrics). These three abstract desiderata are not necessarily consistent, as demonstrated by impossibility results in the literature. 

\subsection{Fairness measures}
Definitions of fairness are diverse across the FML literature. Here we briefly review the main categories of classical fairness (for machine learning) before presenting a quantum analogue of quantum fairness. Following \cite{caton_fairness_2020}, fairness definitions can be categorised into parity fairness, confusion-matrix based fairness, calibration fairness.  
\begin{enumerate}
    \item \textit{Parity metrics.} In a classification context, parity metrics compare parity of positive rates $P(Y=1)$ across different groups. These include: (a) \textit{statistical parity}, namely equal (and thus) probability of positive classification $P(Y=1|g_i)=P(Y=1|g_j)$ without considering group differences; (b) \textit{disparate impact}, the probability of being classified with a positive label, generally considering the ratio between groups i.e. $P(Y=1|g_i)/P(Y=1|g_j)$ where fairness requires the ratio above a threshold (e.g. 80\%).
    %===========
\item \textit{Confusion-matrix based metrics.} Confusion-matrix based metrics consider other typical measures such as true positive rate (TPR), true negative rate (TNR), false positive rate (FPR) and false negative rate (FNR), providing a richer analysis of fairness and discrimination. The key measures include: (a) \textit{equal opportunity}, where TPR is the same across all groups  $P(Y=1|y=1,g_i)=P(Y=1|y=1,g_j)$; (b) \textit{equalised odds} (conditional procedure accuracy equality) where FPR are considered as well, mandating equal probability of false positive and true positive rates conditional upon groups $g_i,g_j$; (c) \textit{overall accuracy equality} concerns the overall percentage of correct (true positives TP and true negatives TN) with $(TN + TP)/(P+N)$, where fairness measures \cite{berk_fairness_2018} are relativised so as to require equal accuracy within groups; (d) \textit{conditional use accuracy equality} assess the predictive values by demanding the probabilities of (positive/negative) classification given positive/negative prediction be equal, that is $P(y=1|Y=1,g_i) = p(y=1|Y=1,g_j)$ and similarly for $y=0$; (e) \textit{treatment equality}, the ratio of FNR to FPR conditioned on each group should be equal; (f) \textit{equalising disincentives}, which requires equality of the \textit{difference} between TPR and FPR conditioned upon groups; and (g) \textit{conditional equal opportunity}, applicable where data is drawn from a distribution $y \sim \mathcal{D}$, such fairness measure require equal probability of prediction above some threshold $\tau$ conditioned on groups, $y$ being below $\tau$ and the data satisfying some attribute $A=A$ (of form $P(\hat{y} > \tau| g_i, y< \tau, A=a)$.
%=====
\item \textit{Calibration.} A third approach compares the predicted probability score $S$ into account. Such approaches include: (a) \textit{test fairness/calibration}, where for different group members, the same predicted probability score should yield the same probability of actually being classified, that is $P(y=1|S=s,g_i)=P(y=1|S=s,g_j)$; (b) \textit{Well calibration}, the same as test fairness calibration but where the probability of classification must equal $s$ itself.
\item \textit{Score-based metrics.} Finally, score-based metrics based upon \textit{expectation} values are also a feature of the literature: (a) \textit{positive/negative class balance}, where fairness is characterised as equality of expected predicted score which should be equal conditioned on predicted score and group; and (b) \textit{Bayesian fairness} measures.
\end{enumerate}
Such definitions of fairness are commonly based on equivalences or equality across probability distributions, with differences arising in what such probabilities are conditioned upon. In the quantum context, measurement results in a probabilistic distribution over (eigenvalues) over chosen measurement operators. These in turn are abstracted as the quantum state $\ketpsi$ residing in a subspace of Hilbert space $\ketpsi \in \mathcal{H}_i \subseteq \mathcal{H}$. Fairness criteria in quantum fair machine learning must be reflected in an available set of measurement operators (or POVM) that \textit{realises} such fairness criteria. This motivates our first definition of \textit{quantum fairness}.\\
\\
\textit{Definition 1: Quantum Fairness}. Given a suitable POVM $\{E_m\} = \{M^\dagger_m M_m\}$, vectors in $\mathcal{H}$ and quantum state $\ketpsi \in \mathcal{H}$ (i.e. $\rho$), the POVM partitions $\mathcal{H}$ (and states) into (possibly disjoint) subspaces $\mathcal{H}_m \in \mathcal{H}_m$. A state $\ketpsi$ satisfies quantum fairness with respect to operators $E_m$ that partition the Hilbert space if $\ketpsi$ is equally likely to reside in each subspace $\mathcal{H}_m$, that is:
\begin{align}
    \braket{\psi| M_m^\dagger M_m | \psi} &= \braket{\psi| M_n^\dagger M_n | \psi} \quad m \neq n\\
    \text{tr}(\rho M_m) &= \text{tr}(\rho M_n)
\end{align} 
Note that we assume that $\rho = \sum_i \rho_i$ where $\rho_i$ encode individual data in quantum states. This general definition of fairness stipulates that $\ketpsi$ has an equal probability of residing within each subspace $\mathcal{H}_m$. Meeting a classification $m$ is equivalent to $\ketpsi$ residing in the partition $\mathcal{H}_m$ associated with the measurement POVM operator $E_m$ that yields measurement (classification) $m$.  

Conditional probabilities, where for example parity (fairness) across multiple classifications, of the type articulated above, become reframed as sequences of measurements on $\ketpsi$ by operators. For example, elementary statistical parity among $n$ groups with regard to a classifiers $\{G_i\},\{Y_k\}$ with classification outcomes (which may be binary or multivariate) $g_i$ and $y_k$ respectively would first require a measurement $G_i$ whose outcome $g_i$ was interpretable as membership of group $\ketpsi \in \mathcal{H}_{g_i}$, followed by a measurement operator $Y_k$ whose realisation $y_k$ indicated classification. The outcome of such measurements would give rise to joint probability distributions (as in the classical case) allowing assessment of whether the multi-classification fairness criteria have been met. However, note that, uniquely in the quantum case as distinct from the classical case, the commutation relations of operators (and thus their order of application) matters: if the operators do commute $[G_i,Y_k]=0$ commute, order of measurement will not matter, while if $[G_i,Y_k]\neq 0$ it will. In the latter case, this means the joint distribution obtained  will differ depending on whether $G_i$ or $Y_k$ is applied first. In this paper, for simplicity, we assume commutativity of measurement operators. This definition of fairness is that we apply in Part III below.

\subsection{Individual and counterfactual fairness.}Extensive research has been undertaken into quantitative measures of individual fairness for several decades. Among these include: (a) \textit{Lipschitz conditioned fairness} \cite{dwork_fairness_2012} imposing individual statistical parity by requiring the difference (measured via metric distance $d_Y$) between outcomes $f(x_1),f(x_2) \in Y$ (usually expressed in terms of distributions) as a result of some function or algorithm $f: X \to Y$ for similar individuals $x_1,x_2 \in X$ to be a bounded function (e.g. constant multiple $K$) of the metric distance $d_X$ of those individuals themselves, that is $d_Y(f(x_1),f(x_2) \leq K d_X(x_1,x_2)$. The condition is expressed typically as a constraint an optimisation problem; (b) \textit{counterfactual fairness}, where given causal models $(U,V,F)$ ($U$ latent background variables, $V=S \bigcup X$ are observables, $S$ are sensitive variables and $F$ structural equation models), fairness requires that individual outcomes are equivalent if sensitive variables had varied; (c) \textit{generalised entropy} comparing individualised prediction to average prediction accuracy (see \cite{speicher_unified_2018} where fairness measures are drawn from axiomatic approaches in economics). In the quantum setting, outcomes are also probability distributions but available metrics are more limited where metrics must compare the similarity of quantum input states (the quantum analogue of $d_X(x_1,x_2)$) and final output states). Unlike the classical Lipschitz condition for fairness \cite{dwork_fairness_2012}, we cannot measure the same quantum state before state evolution $U\ket{\psi(t=0)}$ at $t=0$ (to determine metric similarity of individual data encoded in quantum states) and then afterwards, because $\ketpsi$ will have collapsed to a post-measurement state (\ref{eqn:postmeasurementstate}). With an identical state preparation procedure, we can measure initially and then rely on the creation of identical states with which to evolve.   This motivates a second definition.
\\
\\
\textit{Definition: Quantum Lipschitz Fairness}.
We are given a set of input states $\rho_i = \rho_i(t=0)$ and unitary quantum algorithm $\mathcal{A}(t)$ evolving the state to output state after time $t$ given by $\rho_i' = \mathcal{A}(t)^\dagger \rho_i \mathcal{A}(t)$. The quantum equivalent of input metrics $d_X$ and output metrics $d_Y$ are quantum metrics $D_X,D_Y$ such as trace distance $D(\rho_i,\rho_j) = \frac{1}{2}\text{tr}|\rho_i - \rho_j|$, which is a way to measure state similarity. The quantum analogue of individual fairness (similar output classification/measurement for similar input classification/measurement) is then expressed as a Lipschitz constraint:
\begin{align}
    D_Y(\rho_i', \rho_j') &\leq K(D_X(\rho_i,\rho_j))
\end{align}
where $D$ is a quantum metric described above, such as trace distance, $0< K \leq 1$ and $\rho' = \mathcal{A}^\dagger \rho \mathcal{A}$ i.e. the state after application of the algorithm $\mathcal{A}$. Alternatively, one can compare the distance between sets of inputs and sets of outputs via an entropy-based measure such as quantum relative entropy above such that the quantum Liptschitz condition becomes:
\begin{align}
    |\text{tr}(\rho_i \log \rho_i) - \text{tr}(\rho_i \log \rho_j)| &\leq K|\text{tr}(\rho'_i \log \rho'_i) - \text{tr}(\rho'_i \log \rho'_j)|
\end{align}
A third alternative is to specify two POVMs (i) $\{ S_s \}$ which measure input states after preparation but before application of $\mathcal{A}$ and (ii) $\{ E_m \}$ which measures outputs and compare probability distributions for states $\rho_i,\rho_j$ and $\rho_i',\rho_j'$ with respect to those POVMs. For convenience, the POVMs give rise to a distribution over output measurements of both inputs and outputs (one must repeatedly prepare identical input states for this process). In each case, one must select an appropriate quantum metric to compare distances between input states $\rho_i$ and output states $\rho_i'$.

\subsection{Mitigating unfairness}
Typical techniques within FML literature to mitigate or remedy unfairness are varied. One of the key differences between classical and quantum FML lies in the different mitigation strategies that can be adopted. We explore these more fully below (in the case of binary classification), including preprocessing, model processing and post-processing of outcomes to mitigate biases. Typical methods include the following. 
\begin{enumerate}
    \item \textit{Blinding.} Firstly, (a) \textit{blinding}, where classifiers (and algorithms) are `blind' to protected attributes which are not direct inputs or features in the computation, either via `immunity' (against/to sensitive variables) or omission (which can reduce model accuracy). One challenge is that proxies for sensitive variables can remain (or be reconstructed), leading to potential increases in bias or concealment of discrimination. The inability access information about quantum processes directly will make it difficult to apply similar techniques in the quantum setting. Secondly, (b) \textit{causal methods} in which causal models (such as graphical or probabilistic models) between sensitive and non-sensitive variables, including adding or varying training data in order to meet specific fairness criteria or debiasing, though their effectiveness is varied. Quantum causal models off a quantum analogue that may be applicable.
\item \textit{Sampling and subgroups.} A third approach is in (c) \textit{sampling and subgroup analysis} focused on (i) sampling strategies that alter training data to eliminate unfairness or bias (such as oversampling close to decision-boundaries or thinning out of data away from such boundaries) and (ii) identification of groups or subgroups discriminated against by classifiers, such as via subgroup analysis. Sampling theory and strategies are integral to quantum computing given that one rarely, if ever, has the ability to query the entire Hilbert space $\mathcal{H}$. We assume in this work that we do have such access in a simplified setting, however, understanding how sampling techniques in quantum computing may be used to satisfy fairness criteria is an open an important direction of research.  

\item\textit{Other techniques.} Other mitigation techniques whose applicability quantum settings motivates further research include: (a) \textit{relabelling and perturbation}, the inclusion of perturbative noise in datasets to both improve generalisation \cite{du_quantum_2020}; (b) \textit{re-weighting}, where achieving fairness satisfaction occurs via adjusting quantum amplitudes (we explore this in Part III below); (c) \textit{regularisation and constraint optimisation} there is an extensive literature on constraint optimisation, such as in quantum control contexts \cite{sachkov_control_2009} where, in particular, fairness constraints (and regularisation terms) are encoded in Hamiltonians $H$ in order to steer the controllable part of quantum systems towards desired target states.
\end{enumerate}

\section{Part III: Statistical parity via amplitude amplification}
In this section, we demonstrate how amplitude application using Grover's quantum search algorithms may be used to achieve a measure of statistical parity among groups (or subgroups). This example is designed to illustrate how specifically \textit{quantum} algorithmic techniques may be used to specific FML objectives or optimisations. In reality, as discussed above, statistical parity will likely still mean outcomes remain unfair by some other measures. While the appropriateness of statistical parity as a fairness measure is itself contested within the literature, it remains relevant as a measure in many contexts, including jurisprudential \cite{Burtis_Gelbach_Kobayashi_2017}.

\subsection{Dataset and statistical parity}
We begin by considering a dataset comprising two groups of individual data, binarised into bit strings of length $m$ i.e. $(x_1,..,x_k,...,x_m)$ corresponding to the set $M$ of indicator features $k \in M$ where $|M|=m$. Each bit encodes whether an individual has or does not have a particular attribute i.e. indicator features (for this simple case, we can consider ordinal or continuous variables encoded into indicator variables e.g. an age of 50 would mean the indicator variable for age-range 50 to 69 years would be 1 and all other age interval indicators 0). Let one of the features, $k=s$, indicate that the individual has characteristic (feature) $s$ and is thus a member of a protected class (or subgroup) $S$. Let the dataset $D$, for simplicity, be partitionable into disjoint subsets such that $D=S \oplus G$, where $G$ comprises individuals without the protected attribute $s$. The optimisation problem is to classify members of $S$ and $G$ in order to achieve statistical parity \textit{without} removing protected variables (for reasons discussed above, including decline in accuracy). 

For illustrative purposes, set $m=3$, with the first bit designating membership of the protected class $S$ if $x_1=1$ and the other two features non-protected attributes. The possible combinations of individuals (indexed by $i$) are set-out in Table 1.
% (\ref{table:basisencoding}). 

%=========table of features
\begin{table}
\begin{center}

\begin{tabular}{|c|c|c|c|c|} 
 \hline \hline
 $i$ (index) & $x_1$ (protected) & $x_2$  & $x_3$ & $\ket{x_i^n}$  \\
 \hline \hline
 1 & 1 & 1 & 1 & $\ket{111}$  \\ 
 2 & 1 & 1 & 0 & $\ket{110}$ \\ 
 3 & 1 & 0 & 1 & $\ket{101}$ \\ 
 4 & 1 & 0 & 0 & $\ket{100}$ \\ 
 5 &  0 & 1 & 1 & $\ket{011}$ \\ 
 6 & 0 & 1 & 0 & $\ket{010}$ \\ 
 7 & 0 & 0 & 1 & $\ket{001}$ \\ 
 8 & 0 & 0 & 0 & $\ket{000}$ \\ 
 \hline
\end{tabular}
% \label{table:basisencoding}
\end{center}
\caption{Basis encoding of individuals with features $b_k$ into quantum states $\ket{x^n}$}
\end{table}

We assume that we have a quantum algorithm which we designate (following \cite{brassard_quantum_2000}) as $\mathcal{A}$ acting as a function $f(x)$ designed to solve a particular optimisation problem, such as for example classifying individuals for whom to offer some benefit, like home loans. The motivation for using a quantum (as opposed to classical) algorithm is not important in this example (it might be that a quantum algorithm more feasibly solves the optimisation problem of interest). The raw optimisation problem itself makes no mention of measures of statistical parity or other measures of parity. The sole objective of applying $\mathcal{A}$ is to maximise utility via optimal classification. The specific details of what $\mathcal{A}$ could be and, relevantly for quantum approaches, what type of quantum algorithms might implement such a function are left abstract, though we assume that (a) the quantum algorithm $\mathcal{A}$ uses no quantum measurements and (b) the classical input data to such a problem i.e $(x_1,x_2,x_3)$ are (i) encoded in states $\ket{x_i^n}$ (note: when used in the ket, $x^n_i$ indexes the three-qubit state where $n=3$, not to be confused with the individual qubits which we designate via numerical subscripts $x_1,x_2,x_3$ etc) and then (ii) placed in a superposition state. Examples might include a typical hybrid model of quantum algorithmic learning using quantum variational eigensolvers, for example see \cite{wecker_progress_2015}. 

\subsection{Encoding features}
The first step is to basis-encode the classical data as set-out in Table 1, mapping inputs $(x_1,x_2,x_3) \to c\ket{x_1x_2x_3}, x\in \{0,1\}$. Here $c$ represents the amplitude for the state (which, when a measurement is performed on all three qubits, renders the state as the output with probability $|c|^2$). Upon basis encoding the datasets $D$ into the qubits and placing them into an equal superposition, the state of the system $\ketpsi$ will be:
\begin{align}
    \ketpsi&=\frac{1}{2^n}\sum_{i=1}^{2^n} \ket{x_i^n}
\end{align}
where the amplitude of any state is $1/2^n$. The next step is the application of the quantum optimisation algorithm $\mathcal{A}$, which must be a unitary algorithm without measurement. After applying $\mathcal{A}$, $\ketpsi$ is in the form:
\begin{align}
    \ketpsi&=\sum_{i=1}^{2^n} c_{i}\ket{x_i^n}
\end{align}
where $c_i \in \Cx$ represent the amplitudes for the state $\ket{x_i^n}$ subject to the probability measure constraints that $\sum_i |c_i|^2=1$. The algorithm $\mathcal{A}$ optimises for some objective via adjusting the amplitudes $c_i$ so they are no longer equal. To optimise for the particular objective, each qubit is then measured in the computational basis. If the algorithm $\mathcal{A}$ is indeed optimal, then the most probable measurements obtained from repeating measurements on $\ketpsi$ will be those with the highest amplitudes (thus highest probabilities of being measured), reflecting the optimal choice of states and thus individuals for satisfying the objective in question. Doing so, however, is unlikely to satisfy the fairness constraint of statistical parity between individuals in the protected class (with $x_1=1$) and those not in the protected class (with $x_1=0$).
\\
\\
If we assume the existence of an (oracle) function that indicates, prior to measurement, whether a state $\ket{x_i^n}$ is in a protected class or not (i.e if the first qubit $x_1=1$ or $0$), defined by:
\begin{align}
    \chi(\ket{x_i^n}) = \begin{cases} 
      1 & x_1 = 1 \\
      0 & x_1 = 0 
   \end{cases}
\end{align}
then to achieve statistical parity, we want to approximately equalise the conditional probability of measuring states in protected subclass $(x_1=1)$ and those not in the protected subclass $(x_1=0)$, that is: 
\begin{align}
    Pr(\ketpsi|x_1=0) \approx Pr(\ketpsi|x_1=1) 
    \label{eqn:statisticalparity}
\end{align}
One approach is to simply randomly equalise outcomes via post-processing, but this ignores the fact that certain combinations of states with $x_1=1$ will minimise the loss function embedded in $\mathcal{A}$ than others (that is, certain choices of states with $x_1=1$ will be more optimal than others). Our problem is thus a typical constrained optimisation problem, in this case in the context of some ethically mandated fairness criteria (here, approximate statistical parity). We assume that the optimisation algorithm $\mathcal{A}$ has run in a way that has set amplitudes to satisfy the optimisation problem disregarding fairness. We then apply amplitude amplification to evolve initial states with $x_1=1$ in $\mathcal{H}$ in a way that minimises the objective loss (in relation to $\mathcal{A}$) by comparison with alternative means of achieving statistical parity, such as disregarding $x_1$ and randomly equalising outcomes between the two classifications, which are inconsistent with the objective of optimisation via $\mathcal{A}$. 

\subsection{Amplitude amplification}
To solve the constrained optimisation problem, we apply amplitude amplification methods from \cite{grover_fast_1996, grover_quantum_1997, brassard_quantum_2000,brassard_exact_1997, schuld_supervised_2018, wiebe_quantum_2016} adopting in part the formulation from \cite{nielsen_quantum_2010}. Amplitude amplification is a generalisation of techniques applied to generate a quadratic speedup, of which Grover's algorithm \cite{grover_fast_1996, grover_quantum_1998} (one of the seminal results of theoretical quantum algorithm design) is the most well-known example. The method (set-out in detail in \cite{brassard_quantum_2000}) consists of a series of transformations of quantum data that amplify amplitudes of quantum states that satisfy some optimisation criteria, such as a searching criteria. By amplifying the amplitudes of desirous quantum states, the probability of measuring such states is thereby increased. 
%In a classification context, this means that the probability of classifying such quantum states into a particular class, say a binary 1 or 0 for a given class, is increased. 

For amplitude amplification, we assume the existence of a classifier (an oracle) which disjunctively partitions (classifies) quantum vectors in our Hilbert space $\ketpsi \in \mathcal{H}$ into the direct sum of mutually orthogonal subspaces $\mathcal{H}_1$, the space of quantum states to be classified as 1 and $\mathcal{H}_0$, the space of quantum states to be classified as 0. The partitioning of such states is dependent upon this classifier, which is an oracle $\chi$, in the form a unitary operator. The oracle indicates a solution to our search problem $f(x_1)=1$. Finding $x_1=1$ and can be represented via:
\begin{align}
    \chi&: \mathcal{H} \to \mathcal{H}_1 \\
    \ket{x} &\mapsto (-1)^{f(x)}\ket{x}
    \label{eqn:oracle:nielsen}
\end{align}
which indicates solutions to our classification problem, in this case classifying into $x_1=1,0$, via a phase shift for the desired solution, i.e. phase-shifting (multiplying by -1) all states where, for the first qubit, $x_1=1$. The quantum state $\ketpsi$ belongs to the Hilbert space $\mathcal{H}$. The oracle allows us to partition $\mathcal{H}$ into two subspaces, one for $x_1=1$ and another for $x_0=0$. We denote $\ketpsi$ depending on which subspace it inhabits as follows:
\begin{align}
    \{\ket{\psi_1}\} &= \{\ketpsi| \ketpsi \in \mathcal{H}_1 \}\\
    \{\ket{\psi_0}\} &= \{\ketpsi| \ketpsi \in \mathcal{H}_0 \}
\end{align}
Statistical parity (\ref{eqn:statisticalparity}) is then achieved when there is an equal probability of measuring $\ketpsi \in \mathcal{H}_1$ and $\ketpsi \in \mathcal{H}_0$. That is, in quantum fair machine learning, statistical parity among subgroups is represented by $\ketpsi$ equally likely to inhabit regions of Hilbert classified (and partitioned) by the applicable fairness criteria. This is a result for statistical parity among disjoint subgroups in general. In a typical amplitude amplification setup, such as applying Grover's algorithm, the basis states are initialised in a \textit{uniform} superposition such that their amplitudes are initially identical, this means that it is equiprobable to measure a state in either subspace. Our approach is different. We want to evolve $\ketpsi$ to such an equiprobable state without uniform amplitudes for each state. 
\\
\\
\subsection{Grover's algorithm} It is instructive to understand this process in its simpler form, such as articulated in \cite{nielsen_quantum_2010}. First, we express the space of all states satisfying the search criteria i.e. with $x_1=1$ and all states with $x_1=0$ as two different sums $\sum_{x_1=1}\ket{x^m}$ and $\sum_{x_1=0}\ket{x^m}$. We then define normalised states:
\begin{align}
    \ket{\psi_1}&=\frac{1}{\sqrt{M}}\sum_{x_1=1}\ket{x^m}\\
    \ket{\psi_0}&=\frac{1}{\sqrt{N-M}}\sum_{x_1=0}\ket{x^m}
\end{align}
Our state $\ketpsi$ may be expressed using these two states as a basis such that:
\begin{align}
    \ketpsi&=\sqrt{\frac{M}{N}}\ket{\psi_1} + \sqrt{\frac{N-M}{N}} \ket{\psi_0} 
\end{align}
To achieve the amplification sought via a rotation of $\theta$, we define a unitary operator (so as to preserve quantum coherences and probability measure) $Q(\psi,P) = S_\psi S_\chi$ using the operators:
\begin{align}
    S_\psi &= 2\ket{\psi}\bra{\psi} - \mathbb{I} \\
    S_\chi &= 2O - \mathbb{I}
    %\\
     %       &= -\ket{1}\bra{1} + \ket{0}\bra{0}
     \label{eqn:nielsengrover}
\end{align}
% Here $S_\chi$ is a unitary operator implementing the oracle described above in (\ref{eqn:oracle:nielsen}). 
where $O$ an oracle that partitions the Hilbert space based on the protected attribute (see (\ref{eqn:oracleproject}) below). The operator $S_\chi$ flips phase of states in $\mathcal{H}_1$, i.e. so functions effectively as an oracle. The operator $S_\psi$ flips the phase of the initial state $\ketpsi$. Each operator can be geometrically interpreted as a reflection. The product of these two reflections is a rotation, leading to:
\begin{align}
    \ketpsi = \sin(\theta/2)\ket{\psi_1} +  \cos(\theta/2)\ket{\psi_0}
    \label{eqn:nielsensincos}
\end{align}
where $\cos(\theta/2) = \sqrt{(N-M)/N}$. Applying $Q$ then results in:
\begin{align}
    Q^k\ketpsi=\sin\left( \frac{2k+1}{2} \theta\right) \ket{\psi_1} + \cos\left( \frac{2k+1}{2} \theta\right) \ket{\psi_0}
\end{align}
Apply $Q$ iteratively has the effect of rotating $\ketpsi$ by $\theta$ (geometrically, counterclockwise) so as to increase the amplitude of $\ket{\psi_1}$. The probability of measuring a state in $\mathcal{H}_1$ (i.e. the probability that $\ketpsi \equiv \ket{\psi_1}$) is then:
\begin{align}
    Pr(|\ket{\psi_1}|) = \sin^2\left( \frac{2k+1}{2} \theta\right)
    \label{eqn:nielsengoverprobability}
\end{align}
By applying $Q$ a sufficient number of times, the probabilities of measuring a state in $\mathcal{H}_0$ and $\mathcal{H}_1$ can be approximately equalised so that:
\begin{align}
    \sin^2\left( \frac{2k+1}{2} \theta\right) \approx \cos^2\left( \frac{2k+1}{2} \theta\right)
\end{align}

%=======tikz

% \begin{tikzpicture}
% \draw [color=black!5] (0,0) grid (5,5); 
% \draw (5.5,0) coordinate (a) node[right, below] {$\mathcal{H}_0$}
% -- (0,0) coordinate (b) node[left] %{(0,0)} 
% -- (0,5) coordinate (c) node[left] node[left]{$\mathcal{H}_1$};
% \path  (0,0) coordinate (ad)   -- (25:5cm) coordinate (dd);
% % \draw (ad) -- (dd) coordinate[pos=0.355](c1);
% \coordinate (c2) at ($(c1)!2*2.365 cm!(dd)$); 
% \draw[->,  thick, black]
% let
% \p1=(ad),\p2=(c2),  \n1={{veclen(\x2-\x1,\y2-\y1)}}
% in
% (\x1,\y1) -- (\x2,\y2)
% \pgfextra{\xdef\var{\n1}} ;

% % \fill[black] (\var,0) circle (1.5pt)
% % pic["$\alpha$", draw=black!55, thick, <<->>, angle eccentricity=1.1, angle radius=\var]{angle=a--b--dd} ;

% \draw[->,   black] (0,0)coordinate(o) -- (\var ,0) coordinate (w);
% % \draw[->,   black] (w) -- (0 ,3.5)coordinate(ww) [fill=black] node[left] {$C$};

% \draw  
% pic["$\theta$", draw=black,  angle eccentricity=1.1, angle radius=3cm]{angle=w--o--c2};

% % \draw  [fill=black] (c2) circle [radius=0.15]  node[xshift=-0.35cm, yshift=-0.45cm] {$ww$};
% \end{tikzpicture}

%==========tikz

\begin{figure}
    \centering
    \includegraphics[width=0.8\linewidth]{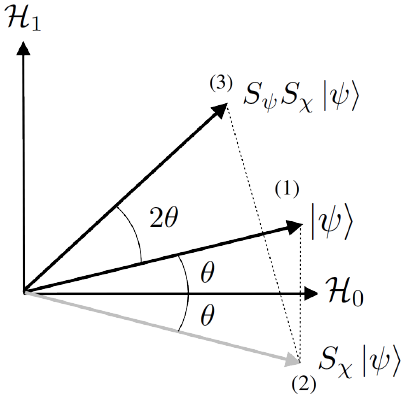}
    \caption{Diagram of amplitude amplification for statistical parity where $\mathcal{H} = \mathcal{H}_0 \oplus \mathcal{H}_1$.  (1) initial state $\ketpsi$ lies closer to $\mathcal{H}_0$, aim is to rotate vector to $\pi/4$ angle between both subspaces. (2) $S_\chi$ operator reflects $\ketpsi$ about the $\mathcal{H}_0$ axis: $\psi \to S_\chi\ketpsi$. (3) Applying $S_{\ketpsi}$ results in $2\theta$ rotation of original $\ketpsi$ vector. Quantum $\epsilon$-statistical parity is achieved after $m = \lfloor \arcsin \sqrt{(0.5-\epsilon} /2\theta - \epsilon \rfloor$ iterations of $Q=S_{\ketpsi}$$S_\chi$.}
    \label{fig:grover}
\end{figure}

The Grover algorithm amplifies amplitudes such that the probability of measuring a state $\ketpsi \in \mathcal{H}_1$ is significantly increased. Our approach is slightly different: in our case, we do not have a uniform superposition, for we have assumed that the algorithm $\mathcal{A}$ has weighted the amplitudes $c_i$ of each basis state (according to some optimisation criteria). Achieving statistical parity across the two groups means obtaining an approximately equal average probability of measuring a state in $\mathcal{H}_1$ or $\mathcal{H}_0$. We can use the same technique, however, to boost amplitudes for $\mathcal{H}_1$ without requiring equiprobability of each state in $\mathcal{H}_1$. We work through our example below in some detail in order to elucidate technical details of how the algorithm works.

\subsection{Statistical parity by amplitude amplification} To amplify amplitudes, we define projection operators which, when applied to $\mathcal{H}$, act to project-out (leave remaining) only those states in $\mathcal{H}_1$ via:
\begin{align}
    P = \ket{1}\bra{1} \otimes \mathbb{I} \otimes \mathbb{I}
\end{align}
As can be seen using density-matrix formalism, this operator acts on the first qubit (leaving the remainder unmeasured, as symbolised by the identity in the tensor product) when applied to $\ketpsi$ results only in states with $x_1=1$.
% For example, acting on a state where $x_1=1$ we have (using $\mathbb{I}=\ket{1}\bra{1} + \ket{0}\bra{0}$):
% \begin{align}
%     P\ket{100} = \ket{1}\bra{1} \otimes \mathbb{I} \otimes \mathbb{I}
% \end{align}
The projection operator is not applied directly to states (for it is not unitary), but rather is used in linear combination with other operators (see below) in order to construct the oracle unitary and the desired amplifying unitary operator. Recalling the basis postulates of measurement described above, if a measurement of $\ketpsi$ is made then the probability of $\ketpsi$ being in a state with $x_1=1$ is given by the amplitude squared:
\begin{align}
    Pr(\ket{\psi_{1}}) = |c_{100}|^2 + |c_{101}|^2 + |c_{111}|^2 + |c_{110}|^2
    \label{eqn:amplitudesum}
\end{align}
The initial probability will not be 50\% in our example because $\ketpsi$ has been subject to $\mathcal{A}$, which has adjusted amplitudes, beforehand. The post-measurement state $\ket{\psi'}$ is:
\begin{align}
    \ket{\psi_1}&=\frac{c_{100}\ket{100} + c_{101}\ket{101}+ c_{111}\ket{111}+c_{110}\ket{110}}{\sqrt{|c_{100}|^2 + |c_{101}|^2 + |c_{111}|^2 + |c_{110}|^2}}
    \label{eqn:psi1state}
\end{align}
The amplification operator $Q(\psi,P) = S_\psi S_\chi$ is then described as the product of the following two operators:
\begin{align}
    S_\psi &= 2\ket{\psi}\bra{\psi} - \mathbb{I} \\
    S_\chi &= 2P - \mathbb{I}
    \label{eqn:oracleproject}
    %\\
     %       &= -\ket{1}\bra{1} + \ket{0}\bra{0}
\end{align}
i.e. equivalent to (\ref{eqn:nielsengrover}) where $O=P$. The unitarity of $Q$ means that $\mathcal{H}$ has an orthonormal basis given by the two eigenvectors of $\mathcal{Q}$ with which we can decompose $\ketpsi$ as: 
% \begin{align}
%     \ket{\psi_\pm}=\frac{1}{\sqrt{2}}\left(\frac{1}{\sqrt{a}} \ket{\psi_1} \pm \frac{i}{\sqrt{1-a}}\ket{\psi_0}  \right)
%     \label{eqn:eigenbasis}
% \end{align}
\begin{align}
    \ket{\psi_\pm}=\frac{1}{\sqrt{2}}\left(\ket{\psi_1} \pm i\ket{\psi_0}  \right)
    \label{eqn:eigenbasis}
\end{align}
Note that in contrast to \cite{brassard_quantum_2000}, we have omitted the prefactors of  ${1}{\sqrt{a}}$ for $\ket{\psi_1}$ and  ${1}{\sqrt{1-a}}$ for $\ket{\psi_0}$ for convenience, including them in the definition of the states $\ket{\psi_1}$ and $\ket{\psi_0}$ respectively.  In that work, such $a$ is designated as the probability of measuring $\ket{\psi_1}$ and $1-a$ the probability of measuring the state $\ket{\psi_0}$. The inclusion of these prefactors in \cite{brassard_quantum_2000} can be thought of as having taken the numerator of (\ref{eqn:psi1state}). To algebraically show the rotational effect of the unitary, we note that the eigenvalues of (\ref{eqn:eigenbasis}) are given by:
\begin{align}
    \lambda_\pm = e^{\pm i2\theta}
\end{align}
using which, we can express $\ketpsi$as:
\begin{align}
    \ketpsi=-\frac{i}{\sqrt{2}}\left(e^{i\theta} \ket{\psi_+} -e^{-i\theta} \ket{\psi_-}  \right)
\end{align}
Which is equivalent in form to (\ref{eqn:nielsensincos}) with $\theta \in [0,\pi/2]$. Applying Q iteratively $m$ times on $\ketpsi$ results in:
\begin{align}
    Q^m \ketpsi &=  -\frac{i}{\sqrt{2}}\left(e^{i(2m+1)\theta} \ket{\psi_+} -e^{-i(2m+1)\theta} \ket{\psi_-}  \right)\\
    &= \sin((2m+1)\theta)\ket{\psi_1} + \cos((2m+1)\theta) \ket{\psi_0}
\end{align}
Each application of $Q$ rotates $\ketpsi$ by angle $2\theta$, thereby projecting more of the state into $\mathcal{H}_1$. The probability that $\ketpsi \in \mathcal{H}_1$ is then $\sin^2((2m+1)\theta)$, consistent with (\ref{eqn:nielsengoverprobability}).  Each rotation by $2\theta$ adjusts the amplitudes for measuring $\ketpsi \in \mathcal{H}_1$ and $\ketpsi \in \mathcal{H}_0$. A diagrammatic representation of this process is set-out in Figure \ref{fig:grover}. The rotation effectively updates the quantum weights (amplitudes) at each step in a unitary. By applying $Q$ a sufficient number of times, we can achieve what we designate as (\textit{quantum}) $\epsilon$\textit{-statistical parity} (namely equal probability within $\epsilon$) such that:
\begin{align}
|\braket{\psi_1|\psi|\psi_1} -\braket{\psi_0|\psi|\psi_0}|&=  \epsilon
    % Pr(|\ket{\psi_1}|) &= Pr(|\ket{\psi_0}|) - \epsilon
    % \sin^2\left( (2m+1) \theta\right) &= \cos^2\left( (2m+1) \theta\right) - \epsilon
\end{align}
which can be expressed as:
\begin{align}
    |0.5 - \epsilon| = \sin^2((2m+1)\theta)
\end{align}
Such statistical parity can be approximated by applying the amplitude amplification operator $Q$ iterative $m$ times where $m$ is given by:
\begin{align}
    m=\left\lfloor \frac{\arcsin\sqrt{(|0.5 - \epsilon|)}}{2\theta} - \theta \right\rfloor
\end{align}
The proportional amplification of the amplitude of states in $\mathcal{H}_1$ is accompanied by a proportional reduction in the amplitude of states in $\mathcal{H}_0$, preserving overall probability measure. Because of the unitarity of $Q$, the adjustment of amplitudes preserves quantum coherence in a way that measuring and classically adjusting probabilities post-measurement would not.
% We setup projection operators that project out states into 
% We define an oracle query $\chi: \Z \to \{0,1\}$ in the form of a unitary operator $U_\omega$ which indicates whether $b_1=1,0$
% In our toy model, we apply this approach as follows. Typical FML problems are cast as constrained optimisation problems,\cite{caton_fairness_2020} whereby the fairness criteria are embedded as constraints on the optimisation problem (see synopsis of methods above). In our example, we assume that the quantum states (and therefore individuals) selected 

\textit{Subgroup parity}. The quantum $\epsilon$-statistical parity method we describe above above can be deployed to achieve statistical parity among any number of $n$ disjoint subgroups so long as there exists a unitary operator $O$ which partitions the Hilbert space $\mathcal{H}$ into a direct sum of subspaces i.e such that:
\begin{align}
    O\mathcal{H} \to \oplus_i^n \mathcal{H}_i = \mathcal{H}_1 \oplus ... \oplus \mathcal{H}_n
\end{align}
In our worked example above, simply setting the projector $P$ to project-out into a particular state of one or more qubits will implement such an operator via $2P - \mathbb{I}$. However, as with classical subgroup parity (see \cite{kearns_empirical_2019}), disjunctively partitioning into one set of subgroups will likely lead to statistical disparity across others. For example, defining the set of projection operators $P$ as:
\begin{align}
    P_{ab}&=\mathbb{I} \otimes \ket{a}\bra{a} \otimes \ket{b}\bra{b} 
\end{align}
where $a,b \in \{ 0,1\}$ will partition $\mathcal{H}$ into for subspaces which can be amplitude amplified into statistical parity. However, there are no guarantees that this retains statistical parity for measurements of the first qubit.

\section{Part IV: Conclusion and future work}
In this first paper on quantum fair machine learning, we have examined foundational characteristics of undertaking fair machine learning involving quantum systems. We have set-out a (by no means exhaustive) comparison between quantum and classical techniques of relevance to FML on quantum systems, elucidating ways in which FML techniques are or are not directly transferable to quantum FML. We have provided a number of quantum analogues for use in quantum FML contexts. As a practical example, we have demonstrated the use of Grover's amplitude amplification algorithm to achieve statistical parity among subgroups and set-out definitions of quantum fairness and quantum Lipschitz-conditioned (individual) fairness. As a new cross-disciplinary field, there are a multitude of open questions and potential research programmes extending from our work including: (a) formalising quantum analogues of existing techniques in FML, (b) exploring QFML in noisy contexts, especially in dissipative open quantum systems, (c) examining how fairness outcomes and computation differs as a result of using quantum-specific resources, such as entanglement and (d) the role of cryptographic and quantum analogues of differential privacy for satisfying fairness criteria for quantum computations.

\begin{acks}
This work is part of the author's research into quantum ethics at the ANU {\it Humanising Machine Intelligence} programme and Centre for Quantum Software and Information at UTS, Sydney. The author thanks colleagues at both institutions for their constructive comments and feedback.
\end{acks}

\bibliographystyle{ACM-Reference-Format}
\bibliography{sample-base}

\appendix

\end{document}